\documentclass[11pt,a4paper]{article}
\usepackage{acl2015}
\usepackage{times}
\usepackage{url}
\usepackage{latexsym}
\usepackage{epsfig}
\usepackage{algorithm}
\usepackage{algorithmic}
\usepackage{tikz} \usetikzlibrary{calc} \tikzset{>=latex}
\usepackage{devanagari}
\usepackage{varwidth}

\DeclareSymbolFont{extraup}{U}{zavm}{m}{n}
\DeclareMathSymbol{\varheart}{\mathalpha}{extraup}{86}

\title{Prepositional Attachment Disambiguation Using Bilingual Parsing and Alignments}

\author{Geetanjali Rakshit$^{\spadesuit\clubsuit\varheart}$ \ \ Sagar Sontakke$^{\clubsuit}$ \ \ Pushpak Bhattacharyya$^{\clubsuit}$ \ \ Gholamreza Haffari$^{\varheart}$ \\ 
  \hspace{15mm} $^{\spadesuit}$IITB-Monash Research Academy, IIT Bombay \hspace{2cm} $^{\varheart}$Monash University \\
  $^{\clubsuit}$ Dept. of Computer Science and Engineering, IIT Bombay \hspace{2cm} Australia\\  
  \hspace{15mm} {\tt geet,sagarsb,pb@cse.iitb.ac.in} \hspace{1cm} {\tt first.last@monash.edu}\\ 
  }

\begin{document}
\maketitle
\begin{abstract}
In this paper, we attempt to solve the problem of Prepositional Phrase (PP) attachments in English. The motivation for the work comes from NLP applications like Machine Translation, for which, getting the correct attachment of prepositions is very crucial. The idea is to correct the PP-attachments for a sentence with the help of alignments from parallel data in another language. The novelty of our work lies in the formulation of the problem into a dual decomposition based algorithm that enforces agreement between the parse trees from two languages as a constraint. Experiments were performed on the English-Hindi language pair and the performance improved by 10\% over the baseline, where the baseline is the attachment predicted by the MSTParser model trained for English.

\end{abstract}

\section{Introduction}
Prepositional Phrase (PP) attachment disambiguation is an important problem in NLP, for it often gives rise to incorrect parse trees . Statistical parsers often predict incorrect attachment for prepositional phrases. For applications like Machine Translation, incorrect PP-attachment leads to serious errors in translation. Several approaches have been proposed to solve this problem. We attempt to tackle this problem for English. English is a syntactically ambiguous language with respect to PP attachments. For example, consider the following sentence where the prepositional phrase \emph{with pockets} may attach either to the verb \emph{washed} or to the noun \emph{jeans}.

\vspace{0.2cm}
\noindent \textbf{Sentence$(1)$:} {I washed the jeans with pockets.}
\vspace{0.2cm}

Below is the correct dependency parse tree (for sentence $1$) where the prepositional phrase \emph{with pockets} is attached to the noun \emph{jeans}.

\begin{figure}
\begin{tikzpicture}
\draw (0,0) node[circle, inner sep=0.8pt, fill=black, label={below:{$I$}}] (A) {};
\draw (1,0) node[circle, inner sep=0.8pt, fill=black, label={below:{$washed$}}] (B) {};    
\draw (2.3,0) node[circle, inner sep=0.8pt, fill=black, label={below:{$the$}}] (C) {};
\draw (3.3,0) node[circle, inner sep=0.8pt, fill=black, label={below:{$jeans$}}] (D) {};  
\draw (4.5,0) node[circle, inner sep=0.8pt, fill=black, label={below:{$with$}}] (E) {};  
\draw (5.8,0) node[circle, inner sep=0.8pt, fill=black, label={below:{$pockets$}}] (F) {};  

\draw[<-] (A) to [bend left=45] (B);
\draw[->] (B) to [bend left=45] (D);
\draw[->] (C) to [bend left=45] (D);
\draw[<-] (D) to [bend left=45] (E);
\draw[<-] (E) to [bend left=45] (F);
\label{dep-parse1}
\end{tikzpicture}
\caption{Dependency Parse Tree for Sentence 1} \label{fig:M1}
\end{figure}
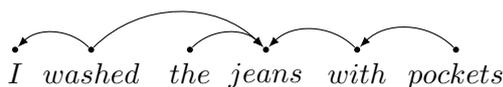

Another possible parse tree for the same sentence could be as shown in Figure \ref{fig:M2}:
\begin{figure}
\begin{tikzpicture}
\draw (0,0) node[circle, inner sep=0.8pt, fill=black, label={below:{$I$}}] (A) {};
\draw (1,0) node[circle, inner sep=0.8pt, fill=black, label={below:{$washed$}}] (B) {};    
\draw (2.3,0) node[circle, inner sep=0.8pt, fill=black, label={below:{$the$}}] (C) {};
\draw (3.3,0) node[circle, inner sep=0.8pt, fill=black, label={below:{$jeans$}}] (D) {};  
\draw (4.5,0) node[circle, inner sep=0.8pt, fill=black, label={below:{$with$}}] (E) {};  
\draw (5.8,0) node[circle, inner sep=0.8pt, fill=black, label={below:{$pockets$}}] (F) {};  

\draw[<-] (A) to [bend left=45] (B);
\draw[->] (B) to [bend left=45] (D);
\draw[->] (C) to [bend left=45] (D);
\draw[<-] (B) to [bend left=45] (E);
\draw[<-] (E) to [bend left=45] (F);
\end{tikzpicture}
\caption{Incorrect Parse Tree for Sentence 1} \label{fig:M2}
\end{figure}
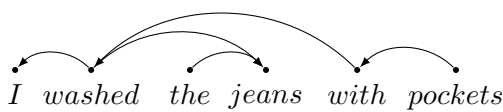
A statistical parser often predicts the PP-attachment incorrectly, and may lead to incorrect parse trees. Let us now look at another sentence.

\vspace{0.2cm}
\noindent \textbf{Sentence$(2)$:} {I washed the jeans with soap.}
\vspace{0.2cm}

\begin{figure}
\begin{tikzpicture}
\draw (0,0) node[circle, inner sep=0.8pt, fill=black, label={below:{$I$}}] (A) {};
\draw (1,0) node[circle, inner sep=0.8pt, fill=black, label={below:{$washed$}}] (B) {};    
\draw (2.3,0) node[circle, inner sep=0.8pt, fill=black, label={below:{$the$}}] (C) {};
\draw (3.3,0) node[circle, inner sep=0.8pt, fill=black, label={below:{$jeans$}}] (D) {};  
\draw (4.5,0) node[circle, inner sep=0.8pt, fill=black, label={below:{$with$}}] (E) {};  
\draw (5.8,0) node[circle, inner sep=0.8pt, fill=black, label={below:{$soap$}}] (F) {};  

\draw[<-] (A) to [bend left=45] (B);
\draw[->] (B) to [bend left=45] (D);
\draw[->] (C) to [bend left=45] (D);
\draw[<-] (B) to [bend left=45] (E);
\draw[<-] (E) to [bend left=45] (F);
\end{tikzpicture}
\caption{Dependency Parse Tree for Sentence 2} \label{fig:M3}
\end{figure}
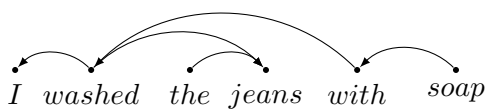

The correct dependency tree for sentence $[2]$ is the following (Figure \ref{fig:M3}), where the prepositional phrase \emph{with soap} attaches to the verb \emph{washed}.

Clearly, there is a case of ambiguity that can be resolved only if the semantics are known. In this case, the fact that \emph{soap} is an aid to the verb \emph{washed} disambiguates its attachment to the verb rather than the noun \emph{jeans}. For correctly translating such an English sentence to another language, the attachments need to be marked correctly.

In this work, we propose a Dual Decomposition (DD) based algorithm for solving the PP attachment problem. We try to disambiguate the PP attachments for English using the corresponding parallel Hindi corpora. Hindi is a syntactically rich language and in most cases exhibits no attachment ambiguities. The use of case markers and the inherent construction of sentences in Hindi make cases of ambiguity rarer. Let us examine how sentences 1 and 2 would look like in Hindi, and if there is a case for ambiguity.

Sentence (3) and sentence (4) are the respective Hindi translations of sentence (1) and (2)). 

\vspace{0.2cm}
\textbf{Sentence $(3)$:} {\dn m\4n\?} {\dn j\?b} {\dn vAlF} {\dn jF\306ws} {\dn DoyF} $|$

\textbf{Sentence $(4)$:} {\dn m\4n\?} {\dn sA\8{b}n} {\dn s\?} {\dn jF\306ws} {\dn DoyF} $|$
\vspace{0.2cm}

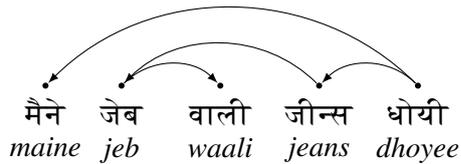
\begin{figure}
\begin{tikzpicture}
\draw (0,0) node[circle, inner sep=0.8pt, fill=black, label={below:{{\dn m\4n\?}}}] (A) {};
\draw (0,-0.5) node[circle, inner sep=0.8pt, label={below:{\emph{maine}}}] (P) {};

\draw (1,0) node[circle, inner sep=0.8pt, fill=black, label={below:{{\dn j\?b}}}] (B) {}; 
\draw (1,-0.5) node[circle, inner sep=0.8pt, label={below:{\emph{jeb}}}] (Q) {}; 

\draw (2.3,0) node[circle, inner sep=0.8pt, fill=black, label={below:{{\dn vAlF}}}] (C) {};
\draw (2.3,-0.5) node[circle, inner sep=0.8pt, label={below:{\emph{waali}}}] (R) {};

\draw (3.6,0) node[circle, inner sep=0.8pt, fill=black, label={below:{{\dn jF\306ws}}}] (D) {};
\draw (3.6,-0.5) node[circle, inner sep=0.8pt, label={below:{\emph{jeans}}}] (S) {};

\draw (4.9,0) node[circle, inner sep=0.8pt, fill=black, label={below:{{\dn DoyF}}}] (E) {};  
\draw (4.9,-0.5) node[circle, inner sep=0.8pt, label={below:\emph{dhoyee}}] (F) {};  

\draw[->] (B) to [bend left=45] (C);
\draw[<-] (A) to [bend left=45] (E);
\draw[<-] (B) to [bend left=45] (D);
\draw[<-] (D) to [bend left=45] (E);
\end{tikzpicture}
\caption{Dependency Parse Tree for Sentence 3} \label{fig:M4}
\end{figure}

In sentence (3), the prepositional phrase \emph{jeb waali} attaches to the noun \emph{jeans} as shown in the figure \ref{fig:M4}.

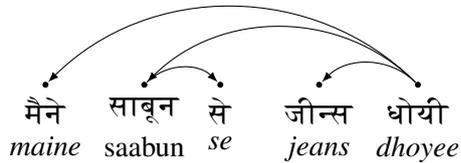
\begin{figure}
\begin{tikzpicture}
\draw (0,0) node[circle, inner sep=0.8pt, fill=black, label={below:{{\dn m\4n\?}}}] (A) {};
\draw (0,-0.5) node[circle, inner sep=0.8pt, label={below:{\emph{maine}}}] (P) {};

\draw (1.3,0) node[circle, inner sep=0.8pt, fill=black, label={below:{{\dn sA\8{b}n}}}] (B) {}; 
\draw (1.3,-0.5) node[circle, inner sep=0.8pt, label={below:{{saabun}}}] (Q) {}; 

\draw (2.3,0) node[circle, inner sep=0.8pt, fill=black, label={below:{{\dn s\?}}}] (C) {};
\draw (2.3,-0.5) node[circle, inner sep=0.8pt,  label={below:{\emph{se}}}] (R) {};

\draw (3.6,0) node[circle, inner sep=0.8pt, fill=black, label={below:{{\dn jF\306ws}}}] (D) {};
\draw (3.6,-0.5) node[circle, inner sep=0.8pt,  label={below:{\emph{jeans}}}] (S) {};

\draw (4.9,0) node[circle, inner sep=0.8pt, fill=black, label={below:{{\dn DoyF}}}] (E) {};  
\draw (4.9,-0.5) node[circle, inner sep=0.8pt, label={below:\emph{dhoyee}}] (F) {};  

\draw[->] (B) to [bend left=45] (C);
\draw[<-] (A) to [bend left=45] (E);
\draw[<-] (D) to [bend left=45] (E);
\draw[<-] (B) to [bend left=45] (E);
\end{tikzpicture}
\caption{Dependency Parse Tree for Sentence 4} \label{fig:M5}
\end{figure}

The parse tree for sentence (4) is shown in figure \ref{fig:M5} , where the prepositional phrase \emph{saabun se} attaches to the verb \emph{dhoyee}.

The case markers \emph{waali} and \emph{se} in the two sentences in Hindi  make the pp-atttachment clear. In our approach, we make use of the parallel Hindi sentences to disambiguate the PP attachments for English sentences. 

The roadmap of the paper is as follows: We discuss the literature and related work for solving the PP-attachment problem in section [\ref{rel-work}]. Section [\ref{our-approach}] describes our approach, and the Dual Decomposition algorithm in detail. The setup, data, and experiments are covered in Section [\ref{exp}]. With Section [\ref{conclusion}], we conclude our work and discuss scope for future work.

\section{Related Work}\label{rel-work}
A number of supervised and unsupervised approaches for solving the PP-attachment problem have been proposed in the literature. \newcite{Ratnaparkhi:94} use a Maximum Entropy Model for solving the PP-attachment decision. \newcite{Schwartz:03} propose an unsupervised approach for solving PP attachment using multilingual aligned data. They transform the data into high-level linguistic representations and use it make reattachment decisions. The intuition is similar to our work, but the approach is entirely different. \newcite{Brill:94} discuss a transformation-based rule derivation method for PP-attachment disambiguation. It is a simple learning algorithm which derives a set of transformation rules from training corpus, which are then used for solving the PP-attachment problem. \newcite{Stetina:97} make use of the semantic dictionary to solve the problem of disambiguating PP attachments. Their work describes use of word sense disambiguation (WSD) for both supervised and unsupervised techniques. Agirre \shortcite{Agirre:08} and Medimi \shortcite{Medimi:07} have used WSD-based strategies in different capacities to solve the problem of PP-attachment. \newcite{Olteanu:05} have attempted to solve the pp-attachment problem as a classification problem of attachment either to the preceding verb or the noun, and have used Support Vector Machines (SVMs) that use complex syntactic and semantic
features. 

\section{Our Approach}\label{our-approach}
We propose a Dual Decomposition based inference algorithm to look at the problem of PP-attachment disambiguation. Dual decomposition, or more generally, Lagrangian Relaxation, is a classical method for combinatorial optimization and has been applied to several inference problems in NLP \cite{Rush:12}. We train two separate parser models for English and Hindi each, using the MSTParser, and make use of these models in the inferencing step. The input to the algorithm is a parallel English-Hindi sentence pair, with its word alignments given. We first obtain the predicted parse trees for the English and Hindi sentences from the respective trained parser models as an initialsiation step. The DD algorithm then tries to enforce agreement between the two parse trees subject to the given alignments. 

\begin{figure}[htpb]
\centering
\includegraphics[scale=0.4]{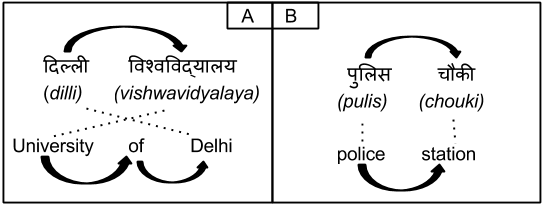}
\caption{Edge Projection from Hindi to English}
\label{fig:pipeline}
\end{figure}

Let us take a closer look at what we mean by agreement between the two parse trees. Essentially, if we have two words in the English sentence denoted by \emph{i} and \emph{i'}, aligned to words \emph{j} and \emph{j'} in the parallel Hindi sentence respectively, we can expect a dependency edge between \emph{i} and \emph{i'} in the English parse tree to correspond to an edge between  \emph{j} and \emph{j'} in the Hindi parse tree, and vice versa. Also, in order to accommodate structural diversity in languages \cite{Smith:06}, we can expect an edge in the parse tree in one language to correspond to more than one edge, or rather, a path, in the other language parse tree. This has been captured in the examples in figures \ref{fig:pipeline}(A) and \ref{fig:pipeline}(B). For an edge in the English parse tree, we term the corresponding edge or path in the Hindi parse tree as the \emph{projection} or \emph{projected path} of the English edge on the Hindi parse tree, and similarly there are projected paths from Hindi to English. For matters of simplicity, we ignore the direction of the edges in the parse trees. The dual decomposition inference algorithm tries to bring the parse trees in the two languages through its constraints.

The problem is formulated as below:
\begin{figure}
\vspace{0.2cm}
\begin{tikzpicture}[node distance = 0cm, auto,->=stealth,point/.style= 
                   {circle,fill=red,minimum size=0pt,inner sep=0pt}]
\tikzstyle{block} = [rectangle, draw,thick,fill=blue!0,
    text centered, minimum height=10em]
\node [block]{%
   \begin{varwidth}{20em}
      $$argmax_{t_e, t_h} \hspace{0.3cm} \bigg( log P_{\theta_{E}} (T_{e}|e) + log P_{\theta_{H}} (T_{h}|h) + $$ $$\sum_{t_{e} \in T_{e}} scr(proj(t_{e}, T_{h})) + \sum_{t_{h} \in T_{h}} scr(proj(t_{h}, T_{e})) \bigg)$$
    \end{varwidth}};
\end{tikzpicture}
\caption{Optimization Problem} \label{objective function}
\end{figure}

In the above formulation, $e$ and $h$ represent a English and Hindi sentence respectively. $T_e$ and $T_h$ are the corresponding parse trees. $\theta_E$ and $\theta_H$ are the model parameters for the edge-factored parser models trained for English and Hindi respectively. $t_e$ represents an edge in the English parse tree $T_e$. $proj(t_{e}, T_{h})$ is a projected path in Hindi parse tree $(T_h)$ for a given English edge $t_e$. The term $scr(proj(t_{e}, T_{h}))$ stands for the score of a projected path in Hindi parse tree $(T_h)$ for a given English edge $t_e$. 

The score of the projected path is calculated as the sum of scores of all edges in the path. Let $\pi_{t_e}$ denote the projected path on sentence \emph{h} in Hindi for the edge $t_e$ in the English parse tree. We assume $scr(\pi_{t_e} = \sum_{a \in \pi_{t_e}} \psi_{t_e}(a)$ where $\psi_{t_e}(a)$ is the score of edge \emph{a} in the projected path $\pi_{t_e}$. In the other direction, $\pi_{t_f}$ and $scr(\pi_{t_f}$ is similarly defined.

To solve this maximization problem in figure \ref{objective function}, we assume one tree to be given and maximize the other and the score of its projected path. The algorithm is described in detail in section \ref{dd-algo}.


\subsection{Dual Decomposition based Algorithm}\label{dd-algo}

We use an iterative \emph{Coordinate Descent} algorithm (Algorithm 1) which calls the  \emph{Project Algorithm} until convergence.
The trees $T_e$ and $T_h$ are initialized by the previously trained parser models for the respective languages. 

\begin{algorithm}
\caption{Coordinate Descent Algorithm}
\label{algo:class_training}
\begin{algorithmic}[1]
\STATE Initialize $T_{e}$ and $T_{h}$ from the MSTParser models 
\STATE \textbf{for} $t = 1$ to $N$
\STATE \hspace{0.5cm} $T_{e}^{+} \leftarrow project(T_{h}, e)$
\STATE \hspace{0.5cm} $T_{h}^{+} \leftarrow project(T_{e}, h)$
\STATE \hspace{0.5cm} \textbf{if} $(T_{e} == T_{e}^{+})$ or $(T_{h} == T_{h}^{+})$
\STATE \hspace{1.0cm} \textbf{break}
\STATE \hspace{0.5cm} \textbf{else}
\STATE \hspace{1.0cm} $T_{e} = T_{e}^{+}$
\STATE \hspace{1.0cm} $T_{h} = T_{h}^{+}$
\STATE \textbf{end for}
\end{algorithmic}
\end{algorithm}

For $N$ iterations, the \emph{project} function returns a parse tree for English which maximizes the agreement between the English and Hindi parse tree when the Hindi parse tree is fixed, and likewise for the Hindi parse tree. The algorithm converges when the trees no longer change,

Let us now look at the \emph{Project} algorithm (Algorithm 2) in detail. It predicts the tree for a sentence in the target language, given the parse tree in the source language, and the word alignments between the parallel sentence.

\begin{algorithm}
\caption{Project Algorithm (tree T, sen S)}
\label{algo:class_training}
\begin{algorithmic}[1]
\REQUIRE A parse tree T (Hindi) and sentence S (English)
\STATE Initialize $\forall t, i, j$  $u_t(i, j)$ = 0
\STATE
\STATE \textbf{for} $t = 1$ to $N$
\STATE \hspace{0.1cm} $Y \leftarrow argmax_{y\in Trees(s)} \hspace{0.1cm} \Big( \sum_{i,j} y(i,j).[\theta (i,j) + r(i,j) - \sum_{t} u_{t}(i,j)] \Big) $
\STATE
\STATE \hspace{0.1cm} \textbf{for} $t \in T$
\STATE \hspace{0.1cm} $\pi_{t} \leftarrow argmax_{\pi \in paths(project \hspace{0.1cm} t \hspace{0.1cm} onto \hspace{0.1cm} S)}$\\ $ \hspace{1.5cm}\sum_{i,j} \pi(i,j) [\varphi_{t}(i,j,S) + u_{t}(i,j)]$
\STATE \hspace{0.1cm} \textbf{end for}
\STATE
\STATE \hspace{0.1cm} \textbf{if} $\forall_{t,i,j}; \pi_{t} = y(i,j)$ then
\STATE \hspace{0.5cm} return $Y$
\STATE \hspace{0.1cm} \textbf{else}
\STATE \hspace{0.5cm} $\forall_{t,i,j}$ \hspace{0.1cm} $ u_{t}(i,j) \leftarrow u_{t}(i,j) - \alpha (\pi_{t}(i,j) - y(i,j))$
\STATE \textbf{end for}
\end{algorithmic}
\end{algorithm}

The lagrangian multipliers are initialized to zero. The best tree in the target language is predicted by the argmax computation in step 4. This maximization involves the parser model parameters $\theta(i, j)$ and the score of the best projected path in the source tree for all edges. $r(i,j)$ denotes the score of the projected path of the edge $y(i, j)$ on the source tree T. In steps 6 and 7, the best projected path for every edge of the source tree is predicted on the target tree using the classifiers described in section \ref{Projected Path Prediction}. The constraints here are that the edges in the projected paths from the classifiers and the predicted trees are in agreement.



\subsection{Projected Path Prediction}
In order to predict the projected path in one language for an edge in the other language, we use a set of two classifiers in a pipeline. Let us recall that we have two nodes in one language with an edge between them, and we are trying to predict the path of the corresponding aligned nodes in the other language. The first classifier predicts the length of the projected path, and the second predicts the predicted path itself, given the path length from the first classifier. Let us look at these classifiers separately. 

The classifier for path length prediction is a set of five binary classifiers, which predict the path length to be 1, 2, 3, 4 or 5. We assume projected path lengths to be no greater than 5. These classifiers are \emph{perceptrons} trained on separate annotated data. The features used were the words and POS tags of the four nodes in the pair of alignments under consideration. 

The classifier for path prediction is a set of four \emph{structured perceptron} classifiers. We train four classifiers to predict the paths of length 2, 3, 4 and 5. These set of classifiers were trained on separate annotated data, and the features used were the same as in the set of classifiers for path length prediction.

\section{Experiments and Results}\label{exp}
A parser model was trained for Hindi using the MSTParser \cite{McDonald:06} by a part of the the Hindi Dependency Treebank data (18708 sentences) from IIIT-Hyderabad \cite{Bhatt:09}. A part of the Penn Treebank (28188 sentences) was used for training an English parser \cite{Marcus:06}. The treebanks were converted to MSTParser format from ConLL format for training. A part of the ILCI English-Hindi Tourism parallel corpus (1500 sentences) was used for training the classifiers.  This corpus was POS-tagged using the Stanford POS Tagger \cite{Toutanova:03} for English and using the Hindi POS Tagger \cite{Reddy:11} from IIIT-Hyderabad for Hindi. It was then automatically annotated with dependency parse trees by the parsers we had trained before English and Hindi. 

For testing, we created a corpus of 100 parallel sentences and their word alignments from the Hindi-English Tourism parallel corpus. We manually annotated the instances of pp-attachment ambiguity.  We examine the prediction for attachment of only these cases. The baseline system used is the attachment predicted by the parser models trained using  the MSTParser. We ran experiments on the test set for iterations 10 to 60, in steps of 10. The outputs from the MSTParser trained model and the DD algorithm were compared against the gold data for English.

Our observations have been tabulated in Table \ref{results}. The MSTParser model was able to correctly disambiguate 54 number of PP-attachments. Our algorithm, however, performed better and  marked 64 number of attachments correctly, in the best case. The baseline accuracy for PP attachment was 54\%. With our approach, we were able to achieve an improvement of 10\% over the baseline.

\begin{table}[h]
\centering
\label{results}
\begin{tabular}{|c|c|c|}
\hline
{\bf Parameter}                                                                         & {\bf \begin{tabular}[c]{@{}c@{}}MST\\ Parser\end{tabular}} & {\bf \begin{tabular}[c]{@{}c@{}}DD\\ Algorithm\end{tabular}} \\ \hline
\begin{tabular}[c]{@{}c@{}}Total Number\\ of PP-attachments\end{tabular}                & 100                                                         & 100                                                           \\ \hline
\begin{tabular}[c]{@{}c@{}}Number of Correctly\\ identified PP-attachments\end{tabular} & 54                                                         & 64                                                           \\ \hline
Accuracy (\%)                                                                           & 54                                                      & 64                                                        \\ \hline
\end{tabular}
\caption{Test Results for English-Hindi}
\end{table}

\begin{figure}[htpb]
\centering
\includegraphics[scale=0.5]{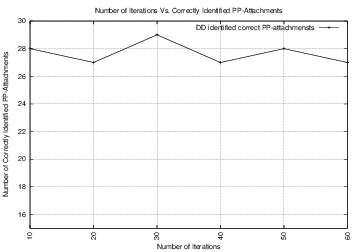}
\caption{Iterations Vs. Correct PP-attachments}
\label{itr}
\end{figure}

We also experimented with the number of iterations to see if the attachment predictions got any better. The observations have been plotted in the graph in figure \ref{itr} . Our algorithm performed best at 30 iterations.

In the event of lack of gold standard data for our experiments, we have used statistical POS taggers for POS tagging the data. Also, for getting word alignments, we have used GIZA++ \cite{Och:03}, which again has scope for errors. These kind of errors may cascade and cause our system to underperform. 

\section{Conclusion and Future Work}\label{conclusion}
We were able to achieve an accuracy of 10\%  over the baseline using our approach. However, in terms of overall dependency parsing and not just with respect to PP-attachment, our system is unable to beat the MSTParser model. However, we need to test our approach on a larger dataset, and across other domains besides Tourism. Besides Hindi, there is also scope for exploring other languages as an aid for pp-attachment disambiguation in English. Our approach could also be used for wh-clause attachment. Since incorrect pp-attachment has a direct consequence on Machine Translation, one interesting analysis could be to use pp-attachments from our system and check for improvement in the quality of translation.


\end{document}